\documentclass[10pt,twocolumn,letterpaper]{article}

\usepackage{iccv}              

\pdfoutput=1
\usepackage{ifpdf}
\usepackage{graphicx}
\usepackage{amsfonts}
\usepackage{amsmath}
\usepackage{kotex}
\definecolor{iccvblue}{rgb}{0.21,0.49,0.74}
\usepackage[pagebackref,breaklinks,colorlinks,allcolors=iccvblue]{hyperref}
\usepackage{caption}
\usepackage{wrapfig}
\usepackage{booktabs}

\usepackage{epsfig}
\usepackage{amssymb}
\usepackage{bbm}
\usepackage{subcaption}
\usepackage{pifont}
\usepackage{longtable}

\usepackage[inline]{enumitem} 
\usepackage[capitalize]{cleveref}
\usepackage{multirow}
\usepackage[makeroom]{cancel}
\usepackage{placeins} 
\usepackage{array}
\newcolumntype{P}[1]{>{\centering\arraybackslash}p{#1}}
\newcolumntype{L}[1]{>{\raggedright\arraybackslash}p{#1}}
\newcolumntype{R}[1]{>{\raggedleft\arraybackslash}p{#1}}

\usepackage{makecell}

\newcommand{\comment}[1]{}

\usepackage{dsfont}

\newcommand{\cmark}{\textcolor{Green}{\ding{51}}}
\newcommand{\xmark}{\textcolor{Red}{\ding{55}}}

\usepackage{algorithm}
\usepackage{algpseudocode}
\crefname{algocf}{alg.}{algs.}
\Crefname{algocf}{Algorithm}{Algorithms}

\floatname{algorithm}{\color{black}Algorithm}

\usepackage{colortbl}

\usepackage{amsmath}

\def\paperID{10} 
\def\confName{ICCV}
\def\confYear{2025}

\title{Leveraging Learned Image Prior for 3D Gaussian Compression}

\author{Seungjoo Shin\\
POSTECH\\
{\tt\small seungjoo.shin@postech.ac.kr}
\and
Jaesik Park\\
Seoul National University\\
{\tt\small jaesik.park@snu.ac.kr}
\and
Sunghyun Cho\\
POSTECH\\
{\tt\small s.cho@postech.ac.kr}
}

\begin{document}
\maketitle

\begin{abstract}
    Compression techniques for 3D Gaussian Splatting (3DGS) have recently achieved considerable success in minimizing storage overhead for 3D Gaussians while preserving high rendering quality. Despite the impressive storage reduction, the lack of learned priors restricts further advances in the rate-distortion trade-off for 3DGS compression tasks. To address this, we introduce a novel 3DGS compression framework that leverages the powerful representational capacity of learned image priors to recover compression-induced quality degradation. Built upon initially compressed Gaussians, our restoration network effectively models the compression artifacts in the image space between degraded and original Gaussians. To enhance the rate-distortion performance, we provide coarse rendering residuals into the restoration network as side information. By leveraging the supervision of restored images, the compressed Gaussians are refined, resulting in a highly compact representation with enhanced rendering performance. Our framework is designed to be compatible with existing Gaussian compression methods, making it broadly applicable across different baselines. Extensive experiments validate the effectiveness of our framework, demonstrating superior rate-distortion performance and outperforming the rendering quality of state-of-the-art 3DGS compression methods while requiring substantially less storage.
\end{abstract}

\section{Introduction}

Recent advances in 3D representations have boosted the performance benchmarks of novel-view synthesis, which aims to model volumetric scenes to synthesize photo-realistic unseen views. One of the representative representations is Neural Radiance Fields (NeRFs)~\cite{mildenhall2021nerf}, which exploits coordinate-based neural networks to enable high-fidelity volumetric rendering. More recently, 3D Gaussian Splatting (3DGS)~\cite{kerbl20233d} has emerged as a prominent representation offering real-time rendering performance with high-quality outputs. Specifically, it introduces 3D Gaussians as point-based primitives parameterized by learnable parameters that determine their shapes and appearances. The remarkable capabilities of 3DGS have inspired numerous subsequent approaches, focusing on improving rendering quality~\cite{yu2024mip,radl2024stopthepop}, dynamic scene modeling~\cite{wu20244d,yang2023real,yang2024deformable,li2024spacetime}, and scene generation~\cite{charatan2024pixelsplat,chen2024mvsplat,xu2024depthsplat}.

Despite its impressive performance, 3DGS suffers from excessive storage requirements, posing significant challenges for real-world applications. This overhead mainly arises from its fine-grained parameterization, where each Gaussian is defined by 59 learnable attributes, including position, covariance, color, opacity, and spherical harmonics coefficients.
To achieve high-fidelity scene representation, 3DGS is initialized from Structure-from-Motion (SfM)~\cite{schonberger2016structure} points and progressively increases the number of Gaussians, often scaling up to millions of Gaussians. While it enables high-quality rendering, it inevitably leads to substantial storage overhead due to the large number of Gaussians and their associated parameters. 
As a result, the storage and memory demands are further amplified, limiting the practicality of 3DGS for scalable deployment, particularly in resource-constrained environments.

To alleviate the storage burden of Gaussians, recent 3DGS approaches propose several compression techniques that aim to minimize the number of Gaussian primitives~\cite{fang2024mini,mallick2024taming}, the redundancy in their attributes~\cite{morgenstern2023compact,niedermayr2024compressed,chen2025fast}, or both objectives~\cite{lee2024compact,chen2024hac,girish2023eagles,shin2025localityaware,fan2023lightgaussian,wang2024contextgs}.
In particular, while attribute-level compression effectively reduces storage overhead, it inevitably introduces information loss, leading to discrepancies between the original Gaussians and their compressed counterparts.
Such distortion in attribute values often results in visible compression artifacts in the rendered images, degrading the overall rendering fidelity.

To mitigate this degradation, we aim to restore the compressed Gaussian attributes to synthesize clean renderings comparable to those obtained from uncompressed Gaussians. Toward this goal, we leverage the strong representational power of learned priors, trained on large-scale data and supported by an effective network architecture. However, constructing learned priors directly in the high-dimensional attribute space is challenging due to the non-trivial distributions of different attribute types and their dependence on scene complexity. To address this, we propose to learn image-level priors from rendered outputs instead of modeling attribute distributions directly. Specifically, our method restores degraded images acquired from compressed attributes by exploiting learned image priors and subsequently employs those restored images as references to refine the degraded Gaussian attributes.

\begin{figure}[t]
\begin{center}
\includegraphics[width=\linewidth]{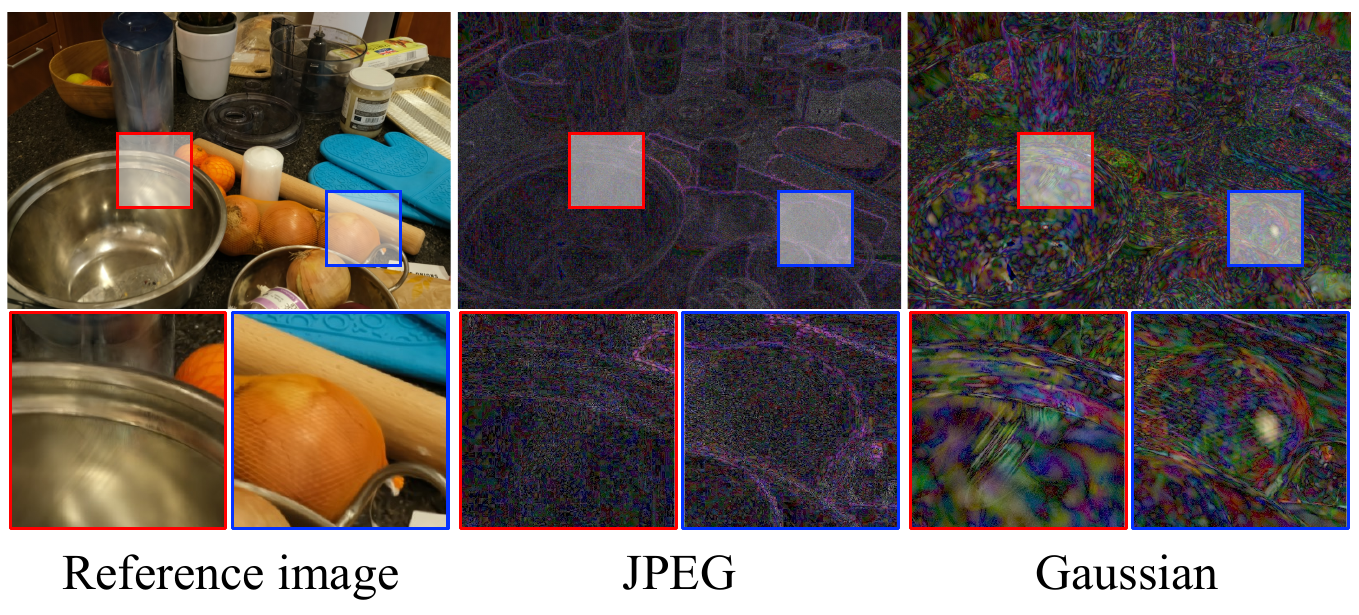}
\end{center}
\vspace{-6mm}
\caption{\textbf{Visualization of compression artifacts in JPEG and Gaussian representation.} We visualize the pixel-wise error maps (MSE) induced by each artifact to highlight the degradation.}
\vspace{-4mm}
\label{fig:degradation}
\end{figure}

In this paper, we propose a novel 3DGS framework that leverages the impressive representational power of learned image priors to restore degraded Gaussians resulting from attribute compression. Given a set of optimized Gaussians, our method first applies a feed-forward Gaussian compression method, FCGS~\cite{chen2025fast}, to build an initial compressed representation with minimal storage and rendering quality. 
We then restore the degraded images rendered from these compressed Gaussians by employing an image restoration network trained to remove visual artifacts introduced by 3DGS compression. Additionally, we provide the coarse residual between the degraded and original renderings to the restoration network as side information. This residual guidance enables the network to capture scene-specific degradation patterns, thereby enhancing the restoration of compressed Gaussians and improving rate-distortion performance. The restored images, containing enhanced visual information, subsequently serve as reference targets to refine the compressed Gaussians for a few additional optimization steps. As a result, experimental evaluations demonstrate that our framework achieves superior rendering quality under comparable storage requirements, compared to baseline methods. Moreover, our approach achieves state-of-the-art (SOTA) rate-distortion performance, outperforming existing 3DGS compression approaches.

We summarize our contributions as follows:
\begin{itemize} [leftmargin=*]
    \item We present a novel 3D Gaussian compression framework that exploits learned image priors to remove Gaussian compression artifacts in rendered images.
    \item Our framework refines compressed Gaussian attributes using restored images to obtain high-quality Gaussians.
    \item Our restoration network successfully models the compression artifacts between the compressed and original Gaussians and further enhances rate-distortion performance by leveraging coarse residuals as side information.
    \item Experimental results show that our framework achieves superior rendering quality under comparable storage requirements and demonstrates SOTA rate-distortion performance over existing 3DGS compression approaches.
\end{itemize}

\section{Related Work}

\subsection{3D Gaussian Splatting and Compression}

3DGS~\cite{kerbl20233d} represents a volumetric scene using a set of anisotropic 3D Gaussians, each of which is a point-based primitive defined by learnable attributes that describe its shape and appearance. By employing tile-based rasterization, this point-based representation enables real-time, high-fidelity novel-view synthesis. Despite its remarkable rendering performance, representing volumetric scenes requires substantial storage space, since a large number of Gaussians are necessary to achieve high-quality rendering, and each Gaussian is associated with several attributes to define intricate shape and appearance. To address this limitation, recent 3DGS approaches have proposed primitive pruning methods~\cite{lee2024compact,fang2024mini,mallick2024taming} and attribute encoding methods~\cite{lee2024compact,morgenstern2023compact,chen2024hac,girish2023eagles,shin2025localityaware, niedermayr2024compressed,fan2023lightgaussian,chen2025fast,wang2024contextgs}. 

\vspace{-4mm}

\paragraph{Primitive-level compression.}
Primitive pruning methods~\cite{kerbl20233d,fang2024mini,mallick2024taming} aim to minimize the number of Gaussian primitives by reducing those with less contribution, using importance score-based pruning. Following the pruning scheme of 3DGS~\cite{kerbl20233d}, several methods~\cite{navaneet2023compact3d,morgenstern2023compact,lu2023scaffold} adopt opacity-based pruning criteria often combined with additional strategies for more efficient pruning. Other approaches exploit carefully designed pruning criteria that account for rendering contribution~\cite{girish2023eagles,fang2024mini}, Gaussian volume~\cite{fan2023lightgaussian}, learnable binary masks~\cite{lee2024compact,wang2024end,chen2024hac,wang2024contextgs,shin2025localityaware}, and multiple components including gradient, per-pixel saliency, and attribute values~\cite{mallick2024taming}. In particular, Mini-Splatting~\cite{fang2024mini} achieves high rendering quality with a minimal number of primitives by leveraging global contribution scores in combination with a deblurring strategy.   

\vspace{-4mm}

\paragraph{Attribute-level compression.}
Attribute compression methods~\cite{lee2024compact,morgenstern2023compact,chen2024hac,girish2023eagles,shin2025localityaware, niedermayr2024compressed,fan2023lightgaussian,chen2025fast,wang2024contextgs} focus on encoding Gaussian attributes in a compact format, which can be efficiently structured in various representations. Several approaches adopt quantization-based techniques that leverage global attribute statistics, including vector quantization~\cite{niedermayr2024compressed,lee2024compact,fan2023lightgaussian,navaneet2023compact3d}, latent quantization~\cite{girish2023eagles}, and image codec~\cite{morgenstern2023compact}. Other methods utilize hash grids to capture the local similarity of attributes~\cite{lee2024compact,shin2025localityaware}. Built upon Scaffold-GS~\cite{lu2023scaffold}, an anchor-based Gaussian representation, HAC~\cite{chen2024hac} and ContextGS~\cite{wang2024contextgs} suggest context modeling to minimize the entropy of anchor features. Despite impressive encoding capability, lacking generalizable priors results in limited compression performance for these per-scene encoding schemes. Recently, FCGS~\cite{chen2025fast} introduced a feed-forward compression network, enabling optimization-free compression within a few seconds. Although the compression network constructs generalizable priors from a large-scale 3DGS dataset, the limited effectiveness of these priors makes it challenging to preserve rendering quality when compressing Gaussian attributes into highly compact storage, due to the unstructured design and high variability of Gaussian representations.


Despite achieving significant storage reductions, most approaches remain limited by their reliance on per-scene information, without exploiting any generalizable priors. To overcome this limitation, we propose a novel 3DGS compression framework that leverages learned image priors to restore compression-induced degradation, thereby improving rate-distortion performance across diverse scenes.


\subsection{Image Restoration}
Image restoration aims to restore a high-quality image from its degraded counterpart. Recently, deep learning methods have significantly improved the performance of image restoration tasks, which include sub-problems such as denoising~\cite{zhang2017beyond,zhang2018ffdnet}, deblurring~\cite{hu2021pyramid}, super-resolution~\cite{zhang2021designing,wang2024exploiting}, and JPEG compression artifact reduction~\cite{yu2016deep,jiang2021towards}. Various neural network architectures, especially U-Net~\cite{ronneberger2015u} variants, are designed to capture specific degradation artifacts and learn generalizable image priors, achieving remarkable performance in image restoration tasks. With the introduction of diverse convolutional neural network (CNN)-based~\cite{dong2015image,yu2016deep,zhang2017beyond,chen2022simple} and transformer-based~\cite{chen2021pre,liang2021swinir,zamir2022restormer} architectures, SOTA performance has continued to be advanced. 

To exploit the strong synthesis capabilities of modern generative models (e.g., GANs or diffusion models), several approaches~\cite{wang2018esrgan,zhang2021designing,wang2021real,wang2024exploiting,lin2024diffbir} incorporate generative priors learned in those models, enabling high-quality image restoration despite severe degradation. Recently, several approaches have adopted generative priors from image restoration models for 3D reconstruction, particularly in tasks such as super-resolution~\cite{feng2024srgs} and few-shot reconstruction~\cite{yang2024gaussianobject,liu20243dgs}. While these methods demonstrate effectiveness in severely degraded reconstruction settings, they might suffer from reduced 3D consistency. In such cases, the generative priors may generate new patterns for each viewpoint, making them less suitable for scenarios where the scene is not severely degraded and geometric consistency must be preserved.

In this manner, we exploit an image restoration model to benefit from the highly expressive learned priors for the 3DGS compression. By employing the restoration network, we successfully recover the degraded images rendered from compressed Gaussians and leverage these restored images to enhance the overall quality of compressed Gaussians. 





\section{Preliminaries: 3D Gaussian Splatting}

Our framework is built upon 3DGS~\cite{kerbl20233d}, which models volumetric scenes using a set of point-based anisotropic Gaussian primitives $\mathcal{G}=(\mathbf{p}, \mathbf{s}, \mathbf{r}, o, \mathbf{k})$. Each Gaussian $\mathcal{G}_i$ is defined by a set of attributes consisting of a position $\mathbf{p}_i \in \mathbb{R}^3$, a scale vector $\mathbf{s}_i \in \mathbb{R}^3$, a quaternion-based rotation vector $\mathbf{r}_i \in \mathbb{R}^4$, an opacity $o_i \in [0, 1]$, and spherical harmonics (SH) coefficients $\mathbf{k}_i \in \mathbb{R}^{3\times(L+1)^2}$, where $L$ is the maximum SH degree. 
The covariance matrix $\Sigma_i \in \mathbb{R}^{3\times3}$ is defined by a rotation matrix derived from $\mathbf{r}_i$, and a scaling matrix derived from $\mathbf{s}_i$. 

For rasterization, 3D Gaussians are projected into 2D space for a given view. The RGB pixel value $\hat{C}(\cdot)$ at view-space coordinate $\mathbf{x} \in \mathbb{R}^2$ is acquired by blending $N$ depth-sorted Gaussians with a view-dependent Gaussian color $\mathbf{c}_i \in \mathbb{R}^3$ derived from SH coefficients $\mathbf{k}_i$ as:
\begin{gather} \label{eq:gaussian}
    \hat{C}(\mathbf{x}) = \sum_{i=1}^{N} \mathbf{c}_i \alpha_i \prod_{j=1}^{i-1} (1 - \alpha_j), \\
    \alpha_i
    = o_i\cdot \exp{(-\frac{1}{2} (\mathbf{x} - \mathbf{p}'_i)^\top \Sigma_i^{'-1} (\mathbf{x} - \mathbf{p}'_i))},
\end{gather}
where $\mathbf{p}'_i$ and $\Sigma_i^{'-1}$ denote projected Gaussian position and covariance, respectively. Initialized from SfM points, the Gaussians $\mathcal{G}$ are optimized to synthesize photo-realistic rendered images for the given set of training viewpoints.

\section{Method}

\begin{figure*}[t]
\begin{center}
\includegraphics[width=\textwidth]{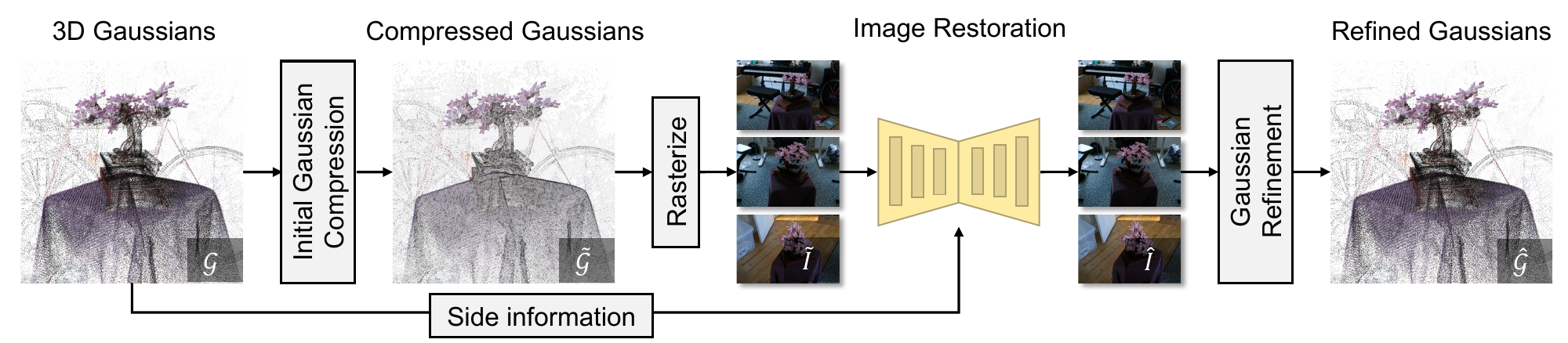} 
\end{center}
\vspace{-4mm}
\caption{\textbf{Illustration of overall pipeline.} Given a set of Gaussians as input, we perform initial compression, followed by an image restoration network that restores the rendered images. Finally, we refine the compressed Gaussians using the restored images to enhance the rendering quality.}
\vspace{-2mm}
\label{fig:framework}
\end{figure*}

\cref{fig:framework} illustrates the overall pipeline of our approach.
Given a set of optimized Gaussians $\mathcal{G}$, we obtain a set of compressed Gaussians $\tilde{\mathcal{G}}$ by applying a compression method (\cref{sec:initial}).
To mitigate the quality degradation, we construct effective side information by computing the coarse residual between the rendered images of the original and compressed Gaussians (\cref{sec:side_info}). Given the degraded renderings and corresponding coarse residuals, we adopt an image restoration network that leverages learned image priors to estimate original rendering images comparable to those produced by the original Gaussians (\cref{sec:restoration}). Finally, these restored rendering images are utilized to refine the compressed Gaussian attributes, resulting in improved rendering fidelity while maintaining compact storage usage (\cref{sec:refine}).

\subsection{Initial Gaussian Compression} \label{sec:initial}
The initial compression step builds compressed Gaussians $\tilde{\mathcal{G}}$ that achieve significant storage savings while introducing rendering distortion as a trade-off. Given a set of optimized Gaussians $\mathcal{G}$, we formulate the compression process as:
\begin{equation} \label{eq:initial}
    \tilde{\mathcal{G}} = \mathcal{H}(\mathcal{G}), \quad
    \tilde{\mathcal{G}}=(\tilde{\mathbf{p}}, \tilde{\mathbf{s}}, \tilde{\mathbf{r}}, \tilde{o}, \tilde{\mathbf{k}}), \quad
    |\tilde{\mathcal{G}}| \leq |\mathcal{G}|,
\end{equation}
where $\mathcal{H}(\cdot)$ denotes a lossy Gaussian compression operation that produces compressed attributes by reducing the number of Gaussians or removing attribute redundancy, at the cost of information loss.
Specifically, we adopt FCGS~\cite{chen2025fast}, a feed-forward 3DGS compression method that achieves remarkable compression performance with minimal computational overhead. Notably, other Gaussian compression methods satisfying \cref{eq:initial} are also compatible with our pipeline as the initial compression module.
The lossy Gaussian compression process inevitably produces information loss, which manifests as noticeable compression artifacts in the rendered outputs, as demonstrated in \cref{fig:degradation}. Mitigating this degradation while preserving storage efficiency remains the core challenge of our framework.

We aim to recover high-quality Gaussians from their compressed counterparts. However, directly modeling the compression-induced distortion is challenging due to the lack of effective priors that can be learned through a specialized network architecture capable of predicting restored Gaussians.
To address this, we reformulate the problem as image-space restoration, where compression artifacts are modeled in the rendered image domain. Specifically, we leverage the rendered images obtained from the original Gaussians $\mathcal{G}$, denoted as $I=\{I_1, ..., I_K\}$, and the degraded images from the compressed Gaussians $\tilde{\mathcal{G}}$, denoted as $\tilde{I}=\{\tilde{I}_1, ..., \tilde{I}_K\}$, to indirectly estimate the attribute residuals using an image restoration network.

\vspace{-1mm}

\subsection{Side Information} \label{sec:side_info}

To achieve more effective restoration for compressed Gaussian renderings, we incorporate side information to support the network to overcome the performance limitations. The degraded images exhibit both compression artifacts with aliasing patterns introduced by the 3DGS rendering process, which differ from typical compression degradations such as JPEG, as shown in \cref{fig:degradation}. Specifically, unlike block-based or frequency-domain artifacts in JPEG, these artifacts often appear as view-dependent, high-frequency distortions that are spatially inconsistent across images. These artifacts result in less effective restoration performance when applying a conventional image restoration framework, indicating the need for additional side information.


\vspace{-2mm}
\paragraph{Coarse rendering residuals.}
To address this challenge, we aim to improve restoration performance by incorporating side information. Specifically, we exploit the residual between the original and degraded images as auxiliary guidance for restoration. However, fully leveraging these high-resolution and high-precision residuals incurs substantial storage overhead. To mitigate this, we compress the residual information $R=I-\tilde{I}$ into coarse residuals $\tilde{R}$ using a compact image codec, JPEG-XL~\cite{alakuijala2019jpeg}, defined as:
\begin{equation}
    \tilde{R} 
    = \mathcal{Q}(R; \lambda_{\textrm{rate}})
    = \mathcal{Q}(I - \tilde{I}; \lambda_{\textrm{rate}}),
\end{equation}
where $\mathcal{Q}(\cdot)$ denotes residual compression operation, and $\lambda_{\textrm{rate}}$ is a hyperparameter that controls the rate-distortion trade-off. By adjusting $\lambda_{\textrm{rate}}$, our framework can flexibly support multiple quality levels, allowing users to balance storage efficiency and restoration quality according to their specific requirements. During the decoding process, we condition the restoration network on the compressed coarse residual $\tilde{R}$ as decoder side information, providing additional guidance to recover high-fidelity renderings from the degraded inputs. Furthermore, we uniformly sample a minimal set of essential training views for side information, leveraging the observation that dataset-provided training views often contain substantial overlap. This allows our framework to operate more efficiently by reducing unnecessary computational and memory overhead.

\begin{figure*}[t]
\begin{center}
\includegraphics[width=\textwidth]{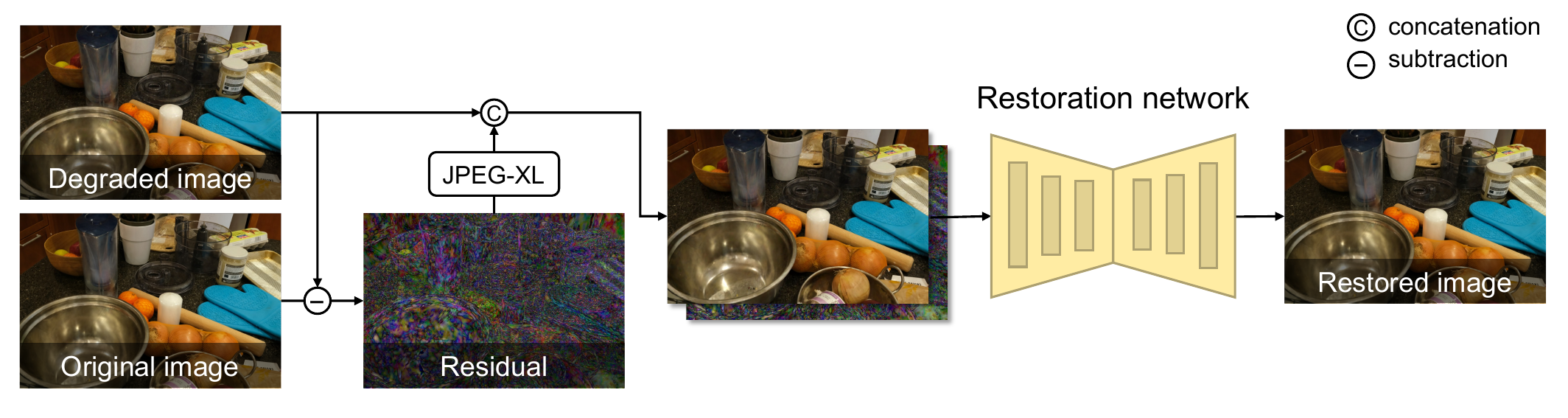}
\end{center}
\vspace{-6mm} 
\caption{\textbf{Illustration of the proposed image restoration framework.} To train the image restoration network for Gaussian compression, we construct a paired dataset consisting of degraded and original images rendered from Gaussians optimized on real-world 3D scenes.
Given a degraded image and the corresponding original image, we first compute their residual. The residual is then quantized and concatenated with the degraded image. This concatenated input is fed into the image restoration network, which is designed to remove Gaussian compression artifacts and recover a high-quality image.}
\vspace{-2mm}
\label{fig:restoration_network}
\end{figure*}

\subsection{Image Restoration} \label{sec:restoration}

\cref{fig:restoration_network} illustrates the proposed restoration framework. Built upon the image restoration network, we opt to model the pixel-level residuals $R=I-\tilde{I}$ between original renderings $I$ and their corresponding degraded renderings $\tilde{I}$ with the support of side information $\tilde{R}$ as:
\begin{equation}
    \hat{I} 
    = \tilde{I} + \hat{R}
    = \tilde{I} + \mathcal{F}(\tilde{I}, \tilde{R}),
\end{equation}
where $\mathcal{F}(\cdot)$ represents the image restoration network, and $\hat{I}$ and $\hat{R}$ denote the restored image and residuals, respectively.
For implementation, we adopt NAFNet~\cite{chen2022simple}, an image restoration architecture that has demonstrated effectiveness in handling various degradation types while maintaining computational efficiency. To incorporate additional side information, we modify the input layer of the network to take a concatenation of the two images, a degraded image $\tilde{I}$, and an auxiliary guidance $\tilde{R}$.

To train the image restoration network, we construct a 3DGS compression artifact dataset consisting of paired images rendered from both original and compressed 3DGS representations. Specifically, we first optimize 3DGS representations on diverse real-world 3D scenes from the DL3DV-10K dataset~\cite{ling2024dl3dv} and subsequently compress the optimized Gaussians using FCGS~\cite{chen2025fast} to obtain compressed Gaussians. We then render images from both the original and compressed Gaussians at multiple training views to generate paired image samples for compression artifact removal. These paired images are finally leveraged to train the image restoration network with the following objectives:
\begin{equation}
    \mathcal{L}_{\textrm{restore}} = \mathcal{L}_1 (R, \hat{R}) + \lambda_{\textrm{LPIPS}}\mathcal{L}_\textrm{LPIPS}(I, \hat{I}),
\end{equation}
where $\mathcal{L}_1$ and $\mathcal{L}_{\textrm{LPIPS}}$ denote $L_1$ loss and LPIPS~\cite{zhang2018unreasonable} loss, and $\lambda_{\textrm{LPIPS}}$ controls the contribution of the perceptual term.


\subsection{Gaussian Refinement} \label{sec:refine}
While the restoration network improves the visual fidelity of degraded renderings, the final objective of our framework is to refine the compressed Gaussian attributes to restore their high rendering ability. We achieve this by re-optimizing the compressed Gaussians $\tilde{\mathcal{G}}$ into refined Gaussians $\hat{\mathcal{G}}$ under the supervision of the restored images $\hat{I}$, effectively propagating the information obtained from the learned image priors. Since the refinement aims to adjust the compressed Gaussians rather than learning from scratch, it is designed to perform with only a few optimization iterations, minimizing computational overhead. The following objective guides the refinement process:
\begin{equation}
    \mathcal{L}_{\textrm{refine}} = (1 - \lambda) \mathcal{L}_{1}(\hat{I}, \hat{I}') + \lambda \mathcal{L}_{\textrm{SSIM}}(\hat{I}, \hat{I}'),
\end{equation}
where $\hat{I}'$ denotes the rendering output from the refined Gaussians $\hat{\mathcal{G}}$, and $\mathcal{L}{1}$ and $\mathcal{L}_{\textrm{SSIM}}$ denote the $L_1$ loss and SSIM loss, respectively. $\lambda$ is a balancing weight. During optimization, we update all Gaussian attributes, whose values are compressed in the initial compression step. This choice can be adjusted based on the specific compression targets of the initial compression method. Also, we preserve the number of Gaussians during the refinement stage.

\begin{figure*}[t]
\begin{center}
\includegraphics[width=0.95\linewidth]{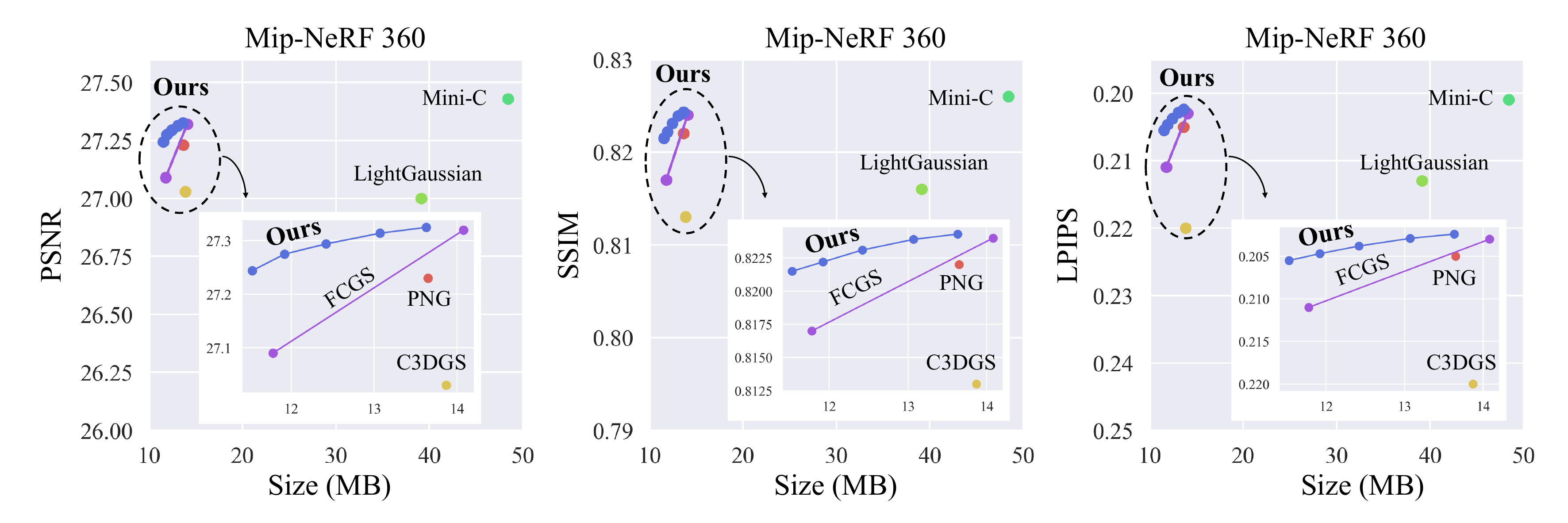}
\end{center}
\vspace{-6mm}
\caption{\textbf{Rate-distortion curves comparing our method with existing post-compression approaches.} We optimize Gaussians following Mini-Splatting~\cite{fang2024mini} and compress them using each post-compression method. Results are averaged over all scenes of Mip-NeRF 360~\cite{barron2022mip}.}
\vspace{-2mm}
\label{fig:quan_mip360}
\end{figure*}

\section{Experiments}

\subsection{Experimental Setting}

\paragraph{Datasets.}

To build a 3DGS compression artifact dataset, we leverage 495 real-world scenes from the DL3DV-10K dataset~\cite{ling2024dl3dv}. As a result, we construct 18,555 image pairs containing visual artifacts introduced by Gaussian compression, which are used to train the image restoration network.
For 3DGS evaluation, we benchmark on two representative novel-view synthesis datasets, Mip-NeRF 360~\cite{barron2022mip}, and Deep Blending~\cite{hedman2018deep}. Specifically, we evaluate nine scenes from Mip-NeRF 360, and two from Deep Blending. Following the evaluation protocol of 3DGS~\cite{kerbl20233d}, we report PSNR, SSIM, and LPIPS to assess rendering quality. In addition, we also evaluate the storage size of each method to compare the compression efficiency.

\vspace{-2mm}

\paragraph{Implementation details.} 
For the image restoration network, we adopt NAFNet~\cite{chen2022simple}, a CNN-based image restoration architecture. We train it for 200K iterations on a single NVIDIA RTX 6000 Ada GPU (48GB). During training, we augment the auxiliary guidance using JPEG compression to simulate diverse quality levels for robust optimization. Following the optimization scheme of Mini-Splatting~\cite{fang2024mini}, we optimize the 3DGS representation for 30K iterations with a minimal number of Gaussians. After optimization, we compress the Gaussian attributes using FCGS~\cite{chen2025fast} and render them to acquire rendering image pairs. We further quantize residual images using a standard image codec, JPEG-XL~\cite{alakuijala2019jpeg}, and employ them as side information for degraded image restoration. To balance storage efficiency and restoration performance, we uniformly sample 40\% of the training views for side information. With supervision from the restored reference images, Gaussian refinement is performed by optimizing the compressed Gaussians for 5K iterations, which takes less than three minutes per scene on a single NVIDIA RTX 6000 Ada GPU (48GB). Furthermore, we provide multiple bitrate variants by adjusting the rate balancing weight $\lambda_{\textrm{rate}}$, which controls the storage cost of the side information. 
Specifically, `Ours-low' uses a smaller $\lambda_{\textrm{rate}}$ to achieve lower bitrate, while `Ours-high' adopts a larger $\lambda_{\textrm{rate}}$ to allocate more storage for improved quality.

\begin{table}[t]
    \centering
    \caption{
        \textbf{Comparison with post-compression approaches.} We evaluate the rendering quality and storage size (MB) on Mip-NeRF 360~\cite{barron2022mip} and Deep Blending~\cite{hedman2018deep}.
    }
    \vspace{-2mm}
    \resizebox{\linewidth}{!}{
    \begin{tabular}{lcccccccc}
        \toprule
            \multirow{2}{*}{Method} &
            \multicolumn{4}{c}{Mip-NeRF 360~\cite{barron2022mip}} &
            \multicolumn{4}{c}{Deep Blending~\cite{hedman2018deep}} \\
            \cmidrule(lr){2-5}  \cmidrule(lr){6-9} 
            & PSNR $\uparrow$ & SSIM $\uparrow$ & LPIPS $\downarrow$  & Size $\downarrow$ 
            & PSNR $\uparrow$ & SSIM $\uparrow$ & LPIPS $\downarrow$  & Size $\downarrow$ \\
        \midrule
        Mini-Splatting~\cite{fang2024mini}
        & 27.39 & 0.827 & 0.196 & 202.9 
        & 30.06 & 0.910 & 0.239 & 138.3 \\ 
        \midrule
        C3DGS~\cite{niedermayr2024compressed}
        & 27.03 & 0.813 & 0.220 & 13.87 
        & 29.80 & 0.903 & 0.257 & 12.10 \\
        LightGaussian~\cite{fan2023lightgaussian}
        & 27.00 & 0.816 & 0.213 & 39.13 
        & 29.56 & 0.900 & 0.253 & 26.48 \\
        Mini-C~\cite{fang2024mini}
        & 27.43 & 0.826 & 0.201 & 48.42 
        & 29.93 & 0.905 & 0.244 & 30.63 \\
        PNG~\cite{ye2025gsplat}
        & 27.23 & 0.822 & 0.205 & 13.65 
        & 29.87 & 0.906 & 0.245 & 8.39 \\
        FCGS-low~\cite{chen2025fast}
        & 27.09 & 0.817 & 0.211 & 11.78 
        & 29.53 & 0.897 & 0.252 & 7.21 \\ 
        FCGS-high~\cite{chen2025fast}
        & 27.32 & 0.824 & 0.203 & 14.08 
        & 29.74 & 0.900 & 0.247 & 8.48 \\
        \midrule
        Ours-low
        & 27.24 & 0.821 & 0.205 & 11.53 
        & 29.96 & 0.906 & 0.248 & 7.21 \\ 
        Ours-high
        & 27.33 & 0.824 & 0.202 & 13.63 
        & 30.01 & 0.907 & 0.247 & 8.14 \\ 
        \bottomrule
    \end{tabular}
    }
    \label{tab:quan_compression}
\vspace{-2mm}
\end{table}

\subsection{Comparison}

To validate the effectiveness of the proposed compression framework, we compare our approach with existing 3DGS compression methods, categorized into \textit{post-compression approaches} and \textit{compact representation approaches}. 

\begin{table}[t]
    \centering
    \caption{\textbf{Comparison with compact representation approaches.} We evaluate rendering quality and storage size (MB) on Mip-NeRF 360~\cite{barron2022mip} and Deep Blending~\cite{hedman2018deep}.}
    \vspace{-2mm}
    \resizebox{\linewidth}{!}{
    \begin{tabular}{lcccccccc}
        \toprule
            \multirow{2}{*}{Method} &
            \multicolumn{4}{c}{Mip-NeRF 360~\cite{barron2022mip}} &
            \multicolumn{4}{c}{Deep Blending~\cite{hedman2018deep}} \\
            \cmidrule(lr){2-5} \cmidrule(lr){6-9} 
            & PSNR $\uparrow$ & SSIM $\uparrow$  & LPIPS $\downarrow$ & Size $\downarrow$ 
            & PSNR $\uparrow$ & SSIM $\uparrow$  & LPIPS $\downarrow$ & Size $\downarrow$ \\
        \midrule
        3DGS~\cite{kerbl20233d}
        & 27.44 & 0.813 & 0.218 & 822.6 
        & 29.48 & 0.900 & 0.246 & 692.5 \\ 
        \midrule
        CompGS~\cite{navaneet2023compact3d} 
        & 27.04 & 0.804 & 0.243 & 22.93 
        & 29.89 & 0.907 & 0.253 & 15.15 \\
        Compact-3DGS~\cite{lee2024compact} 
        & 26.95 & 0.797 & 0.244 & 26.31 
        & 29.71 & 0.901 & 0.257 & 21.75 \\
        EAGLES~\cite{girish2023eagles}
        & 27.10 & 0.807 & 0.234 & 59.49 
        & 29.72 & 0.906 & 0.249 & 54.45 \\
        SOG~\cite{morgenstern2023compact}
        & 27.02 & 0.800 & 0.226 & 43.77 
        & 29.21 & 0.891 & 0.271 & 19.32 \\
        HAC~\cite{chen2024hac}
        & 27.49 & 0.807 & 0.236 & 16.95 
        & 29.99 & 0.902 & 0.268 & 4.51 \\
        LocoGS~\cite{shin2025localityaware}
        & 27.33 & 0.814 & 0.219 & 13.89 
        & 30.06 & 0.904 & 0.249 & 7.64 \\ 
        \midrule
        Ours-low
        & 27.24 & 0.821 & 0.205 & 11.53 
        & 29.96 & 0.906 & 0.248 & 7.21 \\ 
        Ours-high
        & 27.33 & 0.824 & 0.202 & 13.63 
        & 30.01 & 0.907 & 0.247 & 8.14 \\ 
        \bottomrule
    \end{tabular}
    }
    \label{tab:quan_compact}
\vspace{-2mm}
\end{table}

Post-compression approaches aim to reduce storage costs by compressing pre-optimized Gaussians, including C3DGS~\cite{niedermayr2024compressed}, LightGaussian~\cite{fan2023lightgaussian}, Mini-Splatting-C~\cite{fang2024mini}, PNG~\cite{ye2025gsplat}, and FCGS~\cite{chen2025fast}. In particular, C3DGS and LightGaussian require additional optimization steps for fine-tuning, whereas Mini-Splatting-C, PNG, and FCGS are optimization-free methods. For a fair comparison, we equalize the overall optimization budget by performing an additional 5K iterations, matching the refinement setting of our method, before compression for Mini-Splatting-C, PNG, and FCGS. This results in a total of 35K iterations for their initial optimization. In contrast, compact representation approaches directly learn efficient representations from scratch without relying on pre-optimized Gaussians, including SOTA compact Gaussian representations such as CompGS~\cite{navaneet2023compact3d}, Compact-3DGS~\cite{lee2024compact}, EAGLES~\cite{girish2023eagles}, HAC~\cite{chen2024hac}, and LocoGS~\cite{shin2025localityaware}.

\begin{figure*}[t]
\begin{center}
\includegraphics[width=\linewidth]{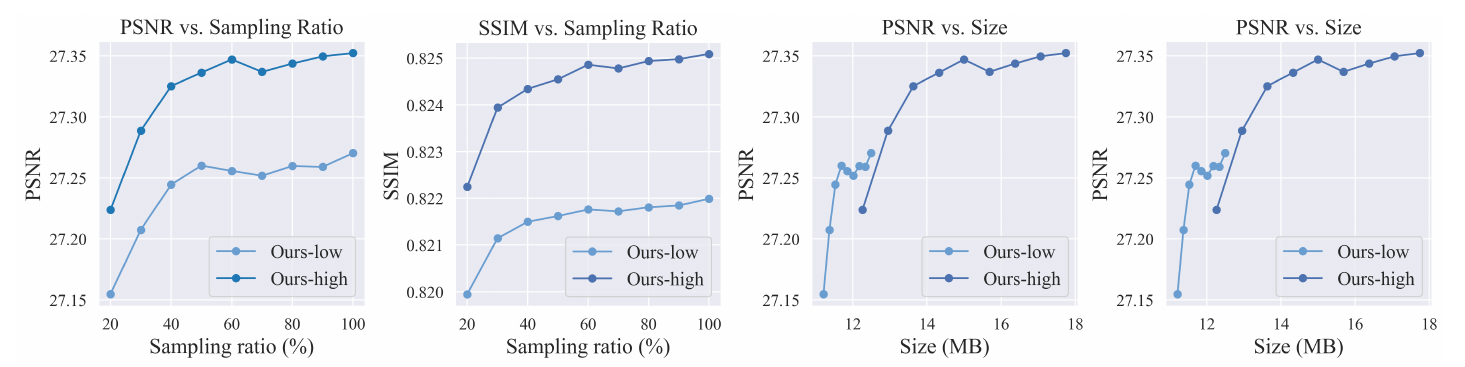}
\end{center}
\vspace{-6mm}
\caption{\textbf{Ablation study on view-sampling.} Results are averaged over all scenes of Mip-NeRF 360~\cite{barron2022mip}.}
\vspace{-4mm}
\label{fig:ablation_sampling}
\end{figure*}

\begin{figure}[ht]
\begin{center}
\includegraphics[width=\linewidth]{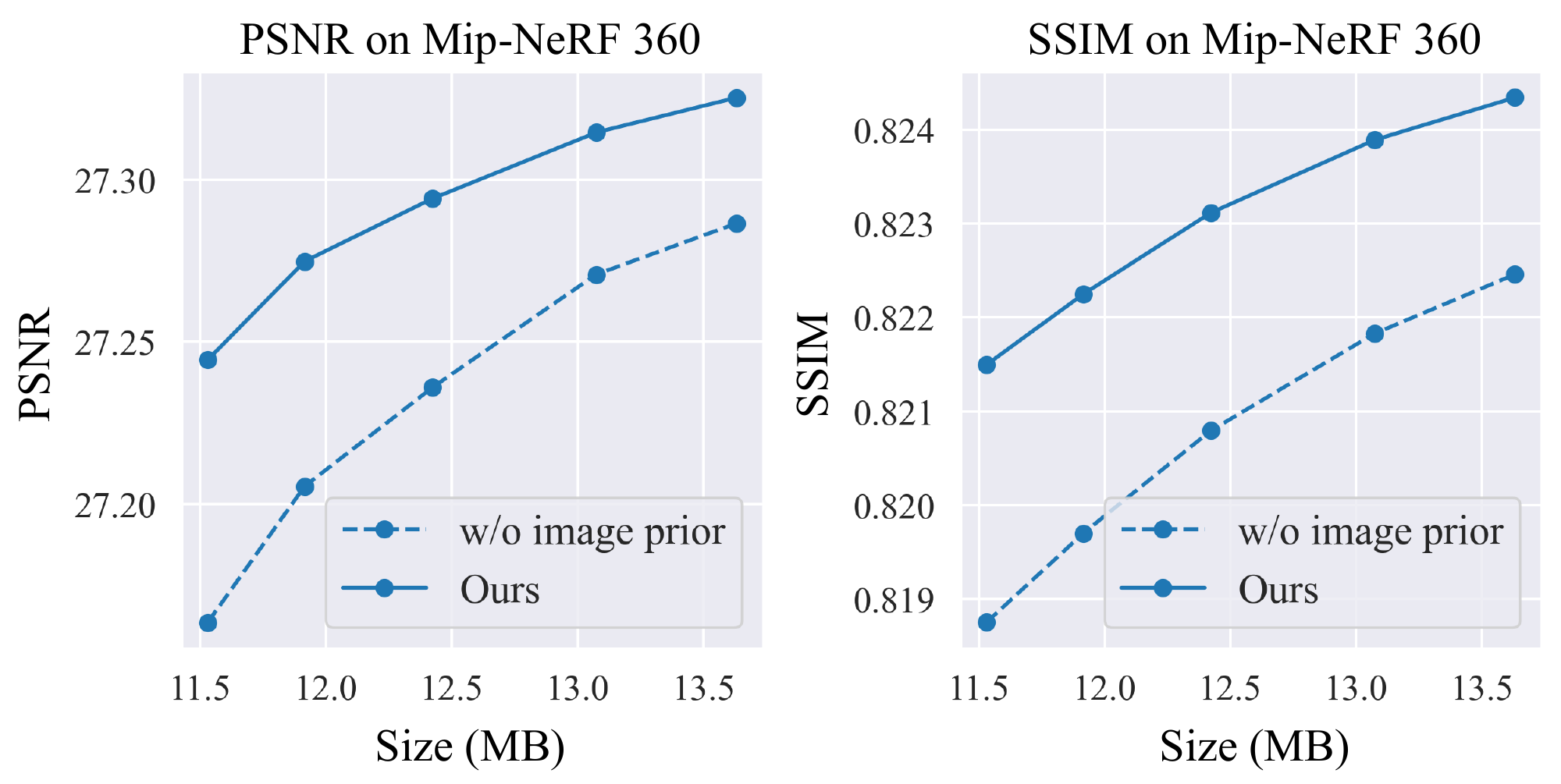}
\end{center}
\vspace{-6mm}
\caption{\textbf{Ablation study on learned image prior.} Results are averaged over all scenes of Mip-NeRF 360~\cite{barron2022mip}.}
\vspace{-4mm}
\label{fig:ablation_prior}
\end{figure}

\vspace{-2mm}

\paragraph{Post-compression approaches.}
\cref{tab:quan_compression} and \cref{fig:quan_mip360} show the quantitative comparison with existing post-compression approaches. Built upon the same initial Gaussians, our compression method demonstrates a superior rate-distortion trade-off, consistently outperforming the rendering quality of baseline methods while achieving significant storage reduction up to 19.2$\times$ compared to the uncompressed representation, Mini-Splatting. Specifically, in the low bitrate setting, our method outperforms the rendering quality of FCGS-low with comparable or even smaller storage consumption.
In the high bitrate setting, our method shows superior rendering performance, except for Mini-Splatting-C, while maintaining the minimal storage cost.
Our method achieves comparable rendering quality to Mini-Splatting-C, which requires up to 3.8$\times$ storage usage. Meanwhile, vector-quantization-based approaches, C3DGS and LightGaussian, suffer from lower rendering performance despite their relatively large storage sizes. These results highlight the remarkable performance of our method as a leading post-compression solution, achieving high-quality rendering and substantial storage savings.

\vspace{-4mm}

\paragraph{Compact representation approaches.}

\cref{tab:quan_compact} presents a quantitative comparison against recent compact Gaussian representations, which optimize Gaussians under compressed designs. Overall, our method achieves superior rendering quality against SOTA compact representations while maintaining the lowest storage overhead, with the exception of HAC on the Deep Blending. Notably, our approach demonstrates considerable perceptual improvements in terms of SSIM and LPIPS, contributing to more faithful and realistic rendering outputs. In contrast, quantization-based methods such as CompGS, Compact-3DGS, EAGLES, and SOG suffer from noticeable degradation in rendering quality despite requiring larger storage sizes. Compared to SOTA compact baselines, HAC, and LocoGS, our method achieves better rate-distortion trade-offs, delivering high-fidelity rendering while preserving minimal storage costs. Specifically, on the Mip-NeRF 360, our method achieves the best performance across all rendering metrics while requiring the smallest storage footprint.

\begin{figure}[t]
\begin{center}
\includegraphics[width=\linewidth]{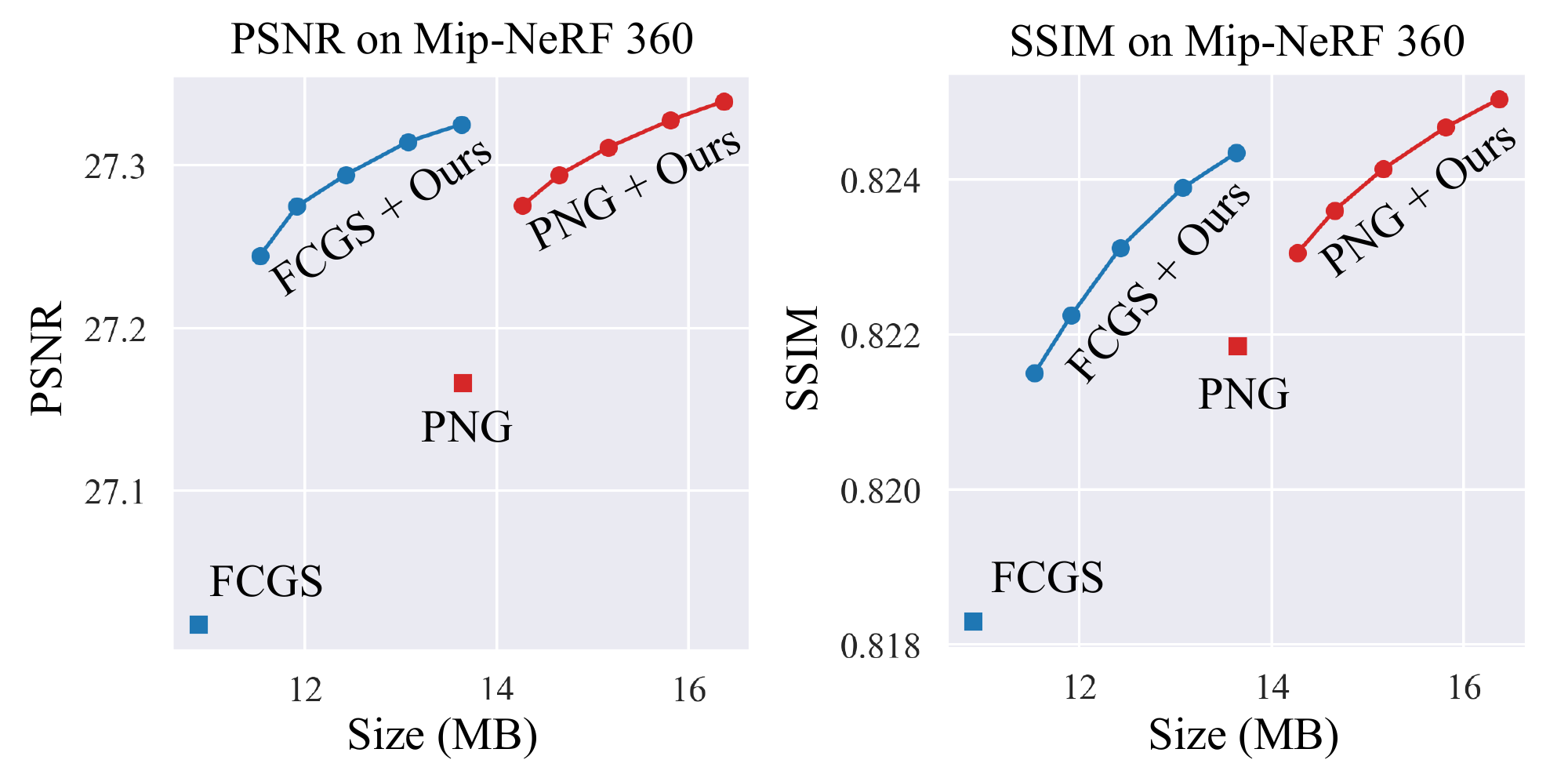}
\end{center}
\vspace{-6mm}
\caption{\textbf{Ablation study on initial compression method.} Results are averaged over all scenes of Mip-NeRF 360~\cite{barron2022mip}.}
\vspace{-4mm}
\label{fig:ablation_base}
\end{figure}

\subsection{Ablation Study}

\paragraph{Learned image prior.}

We investigate the effectiveness of learned image priors for rendering image restoration. To this end, we evaluate the rendering quality of refined Gaussians with and without the use of the trained image restoration model. \cref{fig:ablation_prior} shows the rendering quality of refined Gaussians according to the utilization of learned image priors to the image restoration process. Note that `w/o image prior' refers to the setting where image restoration is performed by simply adding quantized residuals to the degraded image.
In contrast, `Ours' leverages learned image priors embedded in the trained image restoration network to refine the degraded image, as described in \cref{sec:restoration}. As a result, our method achieves noticeable improvements in rendering quality for all evaluation metrics with the support of learned image priors. Overall, these results clearly demonstrate that leveraging learned image priors is crucial for recovering the visual fidelity of compressed Gaussian renderings.

\vspace{-4mm}

\paragraph{View-sampling.}
We also validate the effectiveness of view sampling in balancing storage efficiency and restoration quality. \cref{fig:ablation_sampling} presents the Gaussian refinement performance with respect to the sampling ratio of training views used as side information. Sampling a smaller set of essential views reduces storage consumption by minimizing redundancy while maintaining comparable rendering quality with only minor degradation. Based on this observation, we adopt a 40\% sampling ratio to achieve a favorable balance between storage cost and rendering quality.

\vspace{-2mm}

\begin{table}[t]
    \centering
    \caption{\textbf{Ablation study on side information.} `Image prior' and `Side info.' denote the learned image prior and  side information, respectively. Results are averaged over all scenes of Mip-NeRF 360~\cite{barron2022mip}.}
    \vspace{-2mm}
    \resizebox{\linewidth}{!}{
    \begin{tabular}{cccccc}
        \toprule
         Image prior & Side info. & PSNR $\uparrow$ & SSIM $\uparrow$ & LPIPS $\downarrow$  & Size $\downarrow$ \\
        \midrule
         \xmark & \xmark & 26.98 & 0.816 & 0.212 & 10.89 \\
         \xmark & \cmark & 27.16 & 0.819 & 0.208 & 11.53 \\
         \cmark & \xmark  & 27.00 & 0.817 & 0.211 & 10.89 \\
         \cmark & \cmark & 27.24 & 0.821 & 0.205 & 11.53 \\
        \bottomrule
    \end{tabular}
    }
    \label{tab:ablation_side_info}
\vspace{-2mm}
\end{table}

\paragraph{Side information.}
Moreover, we investigate the effectiveness of side information in enhancing restoration performance. As shown in \cref{fig:degradation}, Gaussian compression artifacts are non-trivial to remove using the restoration network alone, even with a sufficiently large network scale and extensive training dataset. \cref{tab:ablation_side_info} reports the performance of our framework under the difference configuration of the learned image prior and side information. We observe that even a minimal increase in storage size can lead to an evident improvement in rendering quality when accompanied by side information. Also, incorporating side information enables more effective utilization of the learned image prior. Without side information, the learned prior yields only marginal benefits in rendering quality. In contrast, when side information is leveraged, the learned prior provides significantly greater benefits, resulting in enhanced rendering quality.

\vspace{-2mm}

\paragraph{Initial compression method.}
Furthermore, we demonstrate the compatibility of our method with various Gaussian compression approaches. To this end, we incorporate PNG~\cite{ye2025gsplat}, a representative post-compression method, into our compression scheme, as shown in \cref{fig:ablation_base}. Despite the suboptimal rendering quality of the initial compressed results, our method consistently delivers significant improvements, particularly achieving noticeable gains in PSNR, regardless of the underlying compression method. These results indicate that our framework can generalize well and support a wide range of future compression methods.

\section{Conclusion}

In this paper, we introduced a novel 3D Gaussian compression framework that leverages an image restoration network to recover the visual quality of compressed Gaussians. Our approach tackles compression-induced artifacts by reformulating Gaussian restoration as an image-space problem, enabling the use of powerful restoration networks trained on large-scale data. Experimental results demonstrate that our method effectively captures scene-specific degradation patterns and removes compression artifacts, achieving enhanced rendering fidelity and state-of-the-art rate-distortion performance compared to existing 3DGS compression methods, including both post-compression approaches and compact representation approaches.

\vspace{-4mm}

\paragraph{Limitations.}
Despite the impressive restoration performance and broad applicability of our method, it introduces additional computational overhead due to the restoration process. While the additional process of our framework is more efficient compared to existing post-compression methods, it still requires non-negligible computational costs compared to optimization-free approaches.
Moreover, our current design relies on separately quantizing the side information using an external image codec before feeding it into the restoration network. A promising future direction is to improve both computational efficiency and restoration quality by jointly optimizing residual quantization and image restoration within a unified, end-to-end trainable framework.
\section*{Acknowledgments}

This work was supported by the National Research Foundation of Korea (NRF) grant funded by Korea government (Ministry of Education) (No.2022R1A6A1A03052954, Basic Science Research Program), and Institute of Information \& communications Technology Planning \& Evaluation (IITP) grant funded by Korea government (MSIT) (No.RS-2019-II191906, AI Graduate School Program (POSTECH); No.RS-2021-II211343, AI Graduate School Program (Seoul National University)).

{
    \small
    \bibliographystyle{ieeenat_fullname}
    \bibliography{egbib}
}

\end{document}